\PassOptionsToPackage{table}{xcolor}

\documentclass[10pt,twocolumn,letterpaper]{article}

\usepackage[pagenumbers]{cvpr} 
\usepackage[dvipsnames]{xcolor}

\definecolor{cvprblue}{rgb}{0.21,0.49,0.74}
\usepackage[pagebackref,breaklinks,colorlinks,citecolor=cvprblue]{hyperref}

\usepackage[accsupp]{axessibility}
\usepackage{adjustbox}
\usepackage{amssymb}
\usepackage{amsmath}
\usepackage{stmaryrd}
\usepackage{dsfont}
\usepackage{subcaption}
\usepackage{pifont}
\usepackage{wrapfig}
\usepackage{adjustbox}
\usepackage{tabularx}
\usepackage{multirow}
\usepackage{multirow}
\usepackage[dvipsnames]{xcolor}
\usepackage{algorithm}
\usepackage{algpseudocode}

\newcommand{\cmark}{\ding{51}}%
\newcommand{\xmark}{\ding{55}}%
\newcommand{\q}[1]{`#1'}

\def\paperID{4721} 
\def\confName{CVPR}
\def\confYear{2024}
\title{Groupwise Query Specialization and Quality-Aware Multi-Assignment \\ for Transformer-based Visual Relationship Detection}

\newcommand*\samethanks[1][\value{footnote}]{\footnotemark[#1]}
\author{
Jongha Kim\thanks{Equal contribution.} \hspace{0.5cm}
Jihwan Park\samethanks \hspace{0.6cm}
Jinyoung Park\samethanks \\
\hspace{0.1cm}Jinyoung Kim \hspace{0.2cm}
Sehyung Kim \hspace{0.2cm}
Hyunwoo J. Kim\thanks{Corresponding author.} \\
Department of Computer Science and Engineering, Korea University\hspace{0.4cm} \\
\tt\small \{jonghakim, jseven7071, lpmn678, k012100, shkim129, hyunwoojkim\}@korea.ac.kr \vspace{0cm}
}

\begin{document}
\maketitle

\begin{abstract}
Visual Relationship Detection (VRD) has seen significant advancements with Transformer-based architectures recently.
However, we identify two key limitations in a conventional label assignment for training Transformer-based VRD models, which is a process of mapping a ground-truth (GT) to a prediction.
Under the conventional assignment, an \q{unspecialized} query is trained since a query is expected to detect every relation, which makes it difficult for a query to specialize in specific relations.
Furthermore, a query is also insufficiently trained since a GT is assigned only to a single prediction, therefore near-correct or even correct predictions are suppressed by being assigned \q{no relation ($\varnothing$)} as a GT.
To address these issues, we propose Groupwise Query \textbf{Spe}ci\textbf{a}lization and \textbf{Q}uality-Aware Multi-Assignment (SpeaQ).
Groupwise Query Specialization trains a \q{specialized} query by dividing queries and relations into disjoint groups and directing a query in a specific query group solely toward relations in the corresponding relation group.
Quality-Aware Multi-Assignment further facilitates the training by assigning a GT to multiple predictions that are significantly close to a GT in terms of a subject, an object, and the relation in between.
Experimental results and analyses show that SpeaQ effectively trains \q{specialized} queries, which better utilize the capacity of a model, resulting in consistent performance gains with \q{zero} additional inference cost across multiple VRD models and benchmarks.
Code is available at \url{https://github.com/mlvlab/SpeaQ}.
\end{abstract}
\section{Introduction}
Visual Relationship Detection (VRD) is the task of detecting instances (\ie subject, object) and their relation (\ie predicate) given an image, including Scene Graph Generation (SGG) and Human-Object Interaction (HOI) Detection tasks.
The task has a wide range of applications, including image retrieval~\cite{johnson2015image}, visual question answering~\cite{hildebrandt2020scene,shi2019explainable,teney2017graph} and image captioning~\cite{yang2022reformer}. 
Recently, Transformer-based architectures have been increasingly adopted for VRD tasks~\cite{chen2021reformulating,kim2021hotr,li2022sgtr,khandelwal2022iterative}, demonstrating remarkable performances.

To train Transformer-based VRD models, a label assignment is required, which is a process of mapping a ground-truth (GT) to a prediction.
Following DETR~\cite{carion2020end}, the Hungarian matching algorithm~\cite{kuhn1955hungarian} has been a standard of label assignment for Transformer-based VRD models.
However, we observe that queries trained under a standard label assignment are largely \q{unspecialized}, therefore leaving a large portion of a model's capacity underutilized.
To this end, we first identify two major limitations of a standard label assignment that ends up training unspecialized queries.

Firstly, under a standard assignment, a query is trained to detect every relation rather than focusing on a specific relation.
Such multiple roles imposed on a query make it difficult for a query to specialize in a specific role since it provides ambiguous training signals overall.
The long-tailed property of relation distributions of VRD benchmarks even aggravates the problem since unbalanced training signals make it harder for a query to successfully balance between multiple relations.
Secondly, due to a constraint in a standard assignment that a GT can only be assigned to a single prediction, near-correct or even correct predictions are assigned \q{no relation ($\varnothing$)} as a GT, which provides negative signals that suppress the predictions.
For instance, about 45\% of high-quality predictions\footnote{A high-quality prediction is defined as a prediction that is correctly classified and overlaps to the GT on subject and object with IoU over 0.6.} are assigned \q{no relation ($\varnothing$)} as a GT in the case of a model trained on the Visual Genome benchmark.
In sum, an unspecialized query is trained due to multiple roles that defer the specialization of a query and the deficiency in positive training signals under the standard assignment.

To address these limitations, we propose a Groupwise Query \textbf{Spe}ci\textbf{a}lization and \textbf{Q}uality-Aware Multi-Assignment (\textbf{SpeaQ}).
SpeaQ includes two components: Groupwise Query Specialization and Quality-Aware Multi-Assignment.
With groupwise query specialization, a specific target relation group is designated to a query, and a query is trained to only detect relations that belong to a designated relation group.
As a result, a query learns a specialized role instead of struggling to learn to detect every relation due to \q{specific} training signals provided.
Quality-aware multi-assignment further facilitates the training of specialized queries, providing \q{abundant} training signals by assigning a GT to multiple high-quality predictions that are significantly close to GT.

Experimental results demonstrate that SpeaQ be applied to various architectures across Scene Graph Generation (SGG) and Human-Object Interaction (HOI) Detection tasks, resulting in a consistent performance gain.
Due to the specialized queries, SpeaQ covers a wider range of relations where previous models fail (\eg rare relations), while also improving the performance on relations where previous models show decent performance.
As a result, SpeaQ achieves the best performance in both of the two contradicting R@k and mR@k metrics on the VG benchmark~\cite{krishna2017visual}, which are biased toward common and rare relations, respectively.
Notably, such improvements are achieved without any additional post-processing, model parameters, inference cost, or modification in the inference pipeline compared to the baseline.
In sum, our contributions are three-fold:
\begin{itemize}
    \item We introduce a Groupwise Query Specialization, which trains a \q{specialized} query by dividing queries and relations into disjoint groups and directing a query solely toward relations in a corresponding relation group.
    \item We propose a Quality-Aware Multi-Assignment which assigns a GT to multiple predictions considering the triplet-level prediction quality, therefore adaptively providing richer training signals to promising predictions.
    \item Overall, Groupwise Query \textbf{Spe}ci\textbf{a}lization and \textbf{Q}uality-Aware Multi-Assignment (SpeaQ) effectively train a specialized query, which better leverages the model capacity therefore consistently improves performance across multiple VRD models and benchmarks with \textit{zero} additional inference cost.
\end{itemize}
\section{Related Works}
\subsection{Transformers for Visual Relationship Detection}
Visual Relationship Detection (VRD), including Scene Graph Generation (SGG)~\cite{krishna2017visual} and Human-Object Interaction (HOI) Detection~\cite{chao2018learning} is a task of detecting triplets existing in an image, where a triplet consists of instances (\ie subject, object) and a relation between those instances (\ie predicate).
Recently, a line of research developing better Transformer-based~\cite{vaswani2017attention} architectures for VRD tasks has been conducted~\cite{chen2021reformulating,li2022sgtr,khandelwal2022iterative,dong2021visual,kim2021hotr,tamura2021qpic,zhang2021mining,liao2022gen,park2022consistency} following the success of DETR~\cite{carion2020end}.
In this paper, we propose a way to better train Transformer-based VRD models, which can be applied to multiple architectures to better leverage the capacity of those models.

\subsection{Effective training of VRD models}
To mitigate the long-tailed property of VRD benchmarks, multiple learning strategies have been proposed, including data resampling~\cite{li2021bipartite,desai2021learning}, loss re-weighting~\cite{yan2020pcpl,zareian2020bridging,khandelwal2022iterative}, and building class-specific classifiers~\cite{dong2022stacked}.
However, such an approach inevitably results in a loss in common classes since it seeks a trade-off between common and rare classes under the same model capability.
Our work differs from these approaches in that ours enhances the model's capability itself by training a specialized query, therefore improving performance across classes regardless of frequency.
On the other hand, recent works in object detection tried tailoring a label assignment process for detectors, including multiple works actively providing training signals to predictions with low localization costs~\cite{ouyang2022nms,chen2023group,jia2023detrs,hong2022dynamic,wang2021end} on a single object.
Motivated by those works, we introduce an enhanced label assignment strategy for VRD tasks that comprehensively considers a triplet-level localization and classification quality, which is an initiative work exploring the better label assignment strategy for VRD tasks.

\section{Method}
\begin{figure*}[ht!]
\begin{center}
\vspace{-0.5cm}
\includegraphics[width=1.0\textwidth]{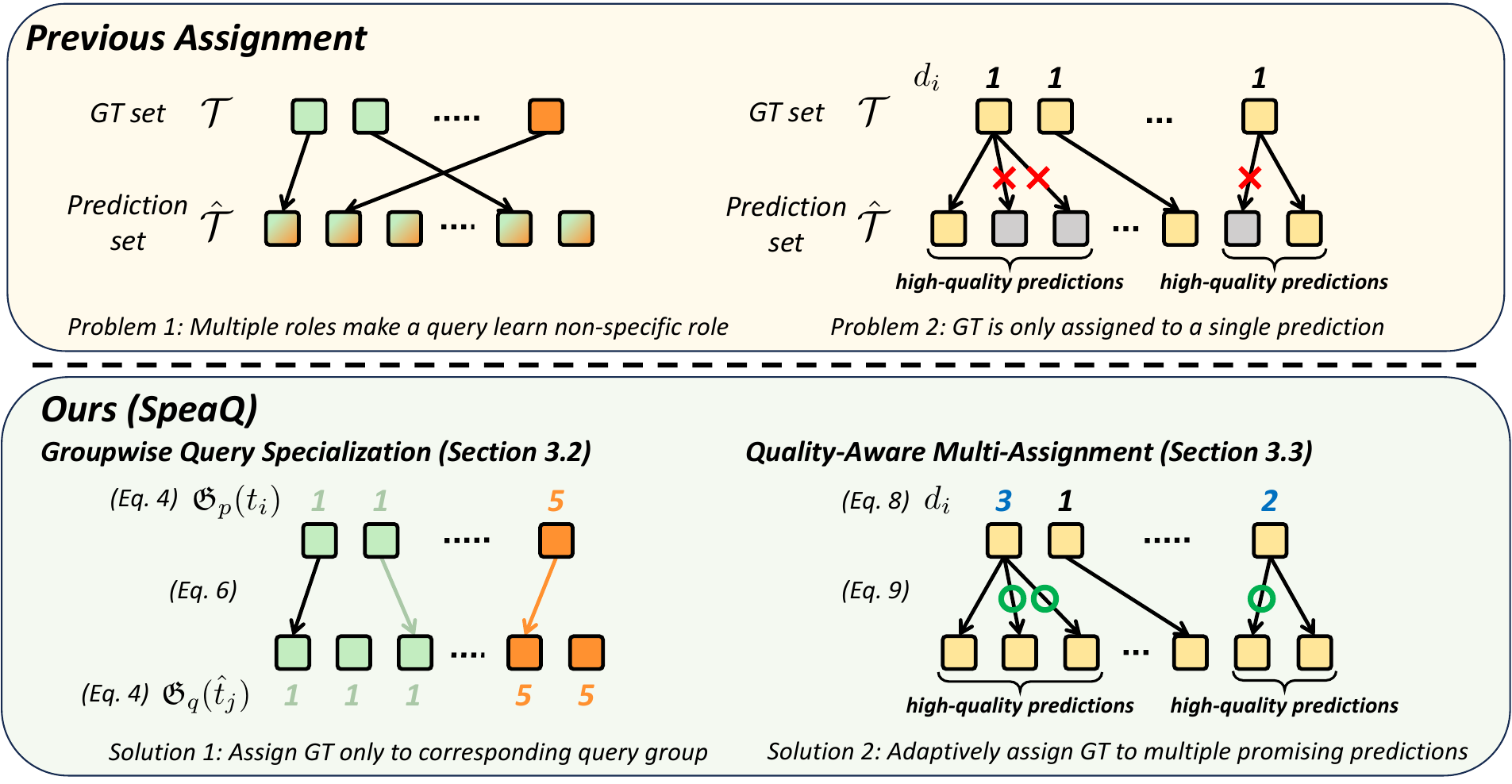}
\end{center}
\vspace{-0.5cm}
\caption{
\textbf{Overview of the proposed SpeaQ.} 
SpeaQ consists of two key components: Groupwise Query Specialization and Quality-Aware Multi-Assignment. 
Groupwise Query Specialization (Sec.~\ref{sec:sec_query_grouping}) divides predicates and queries into disjoint predicate groups and query groups and assigns a GT in a specific predicate group only to a query in the corresponding query group, therefore designating a specialized role to a query. 
Quality-Aware Multi-Assignment (Sec.~\ref{sec:sec_qg_dla}) adaptively assigns a GT to a different number of predictions considering overall prediction quality on a subject, object, and predicate to provide richer training supervision to predictions that are close to a GT. 
}
\label{fig:main_figure}
\vspace{-0.5cm}
\end{figure*}

In this section, we briefly introduce the structure of Transformer-based VRD models and the standard label assignment strategy (Sec.~\ref{sec:prelim}). 
We then propose a Groupwise Query Specialization which directs a query toward a specific predicate group (Sec.~\ref{sec:sec_query_grouping}). 
We also present a Quality-Aware Multi-Assignment which assigns a GT to multiple high-quality predictions considering the triplet-level prediction quality, and the overall pipeline (Sec.~\ref{sec:sec_qg_dla}).
\subsection{Preliminary}
\label{sec:prelim}
\noindent\textbf{Transformer-based visual relationship detection.}
A Visual Relationship Detection (VRD) dataset $\mathcal{D} = \{(\mathcal{I}_i, \mathcal{T}_i)\}^{|\mathcal{D}|}_{i=1}$ consists of pairs of an image $\mathcal{I}_i$ and a corresponding GT set $\mathcal{T}_i$.
A GT set $\mathcal{T}_i = \{t_j = (s_j, p_j, o_j)\}^{|\mathcal{T}_i|}_{j=1}$ is a set of GT triplets $t_j$, where a triplet $t_j = (s_j, p_j, o_j)$ consists of a subject $s_j$, a predicate $p_j$ and an object $o_j$.
Note that the \q{relation} between instances are often termed as \q{predicate} in the context of VRD.
The subject, predicate and object are represented by bounding boxes $b^s_j, b^p_j, b^o_j \in \mathbb{R}^4$ and class labels $c^s_j, c^p_j, c^o_j$.
Note that $\mathcal{T}_i$ includes \q{no relation ($\varnothing$)} label padded to $N_t$ GT labels, so that $|\mathcal{T}_i| = N_q$ holds, where $N_q$ is the number of decoder queries.
Given an image $\mathcal{I}_i$, typical Transformer-based VRD models output the set of predictions $\hat{\mathcal{T}}_i = \{\hat{t}_j = (\hat{s}_j, \hat{p}_j, \hat{o}_j)\}^{N_q}_{j=1}$.
Typical Transformer-based VRD models consist of a CNN backbone and encoder-decoder Transformers.
A conventional CNN backbone network (\eg ResNet~\cite{he2016deep}) first generates a visual feature $\mathcal{F}_i \in \mathbb{R}^{C\times H\times W}$ given an input image $\mathcal{I}_i$.
Then, the visual feature $\mathcal{F}_i$ is fed into a Transformer encoder which outputs an encoded feature $\mathcal{Z}_i \in \mathbb{R}^{C \times HW}$.
Transformer decoders take $\mathcal{Z}_i$ as a feature for cross-attention and transform $\mathcal{Q} = \{q_j\}^{N_q}_{j=1}$, the set of $N_q$ learnable queries into output embeddings.
Finally, output embeddings are translated into final predictions, where the set of whole predictions is denoted as $\hat{\mathcal{T}}_i$.

\noindent\textbf{Label assignment for Transformer-based VRD models.}
Label assignment maps a ground-truth to a prediction to train a Transformer-based VRD model.
For the label assignment in Transformer-based architectures, the Hungarian matching algorithm~\cite{kuhn1955hungarian} that finds a one-to-one assignment between ground-truths and predictions is widely adopted.
Given a GT set $\mathcal{T}_i$ and a prediction set $\hat{\mathcal{T}}_i$, Hungarian matching algorithm finds $\sigma^*_{\text{hungarian}} \in \mathfrak{S}_{N_q}$, the permutation of predictions with the minimal matching cost below:
\begin{equation}
    \sigma^*_{\text{hungarian}} = \underset{\sigma \in \mathfrak{S}_{N_q}}{\mathrm{argmin}}\sum_i^{N_q}\mathcal{H}_{\text{match}}\left(t_i, \hat{t}_{\sigma(i)} \right),
\label{eq:o2o}
\end{equation}
where $\mathcal{H}_{\text{match}}$ is a matching cost between a ground-truth $t_i$ and a prediction $\hat{t}_{\sigma (i)}$ with an index $\sigma(i)$.
In VRD tasks, the overall matching cost $\mathcal{H}_{\text{match}}$ is defined as:
\begin{equation}
\small
\begin{split}
\label{eq:matching_cost}
    &\mathcal{H}_{\text{match}}\left(t_i, \hat{t}_{\sigma(i)} \right) \\
    &= \mathds{1}_{\left\{t_i \neq \varnothing \right\}}
    \begin{bmatrix}
        \mathcal{C}_{s}\left( s_i, \hat{s}_{\sigma(i)} \right) 
        + \mathcal{C}_{p}\left(p_i, \hat{p}_{\sigma(i)} \right)
        + \mathcal{C }_{o}\left(  o_i, \hat{o}_{\sigma(i)} \right)
    \end{bmatrix},
    \end{split}
\end{equation}
where $\mathds{1}$ is an indicator function and $\mathcal{C}_s, \mathcal{C}_p, \mathcal{C}_o$ denote subject, predicate and object matching cost, respectively.
Each matching cost consists of a classification cost (\eg cross-entropy loss) and the sum of localization costs (\eg L1 and generalized IoU loss).
\subsection{Groupwise Query Specialization}
\label{sec:sec_query_grouping}
\paragraph{Frequency-based predicate and query grouping.}
To let a query specialize on specific target predicates, we first divide the set of whole predicate classes into $N_g$ disjoint predicate groups $\{\mathcal{G}^p_i\}_{i=1}^{N_g}$ based on their frequencies.
Predicates with similar frequencies are grouped to ensure a more balanced distribution of frequencies within each predicate group, which helps to avoid optimization difficulties caused by the class imbalance.
Further details about the predicate grouping are provided in the Sec. A and Alg. 1 of the supplementary material.
The set of query $\mathcal{Q}$ with $N_q$ queries is also divided into $N_g$ query groups $\{\mathcal{G}^q_i\}_{i=1}^{N_g}$.
To divide the query set $\mathcal{Q}$ into $N_g$ groups, we propose a \textit{proportional query grouping}, where the number of queries in the $k$'th query group is set proportional to the sum of frequencies of predicates in the $k$'th predicate group in the training set, formulated as:
\begin{equation}
|\mathcal{G}^q_k| \propto \sum_{i}^{|\mathcal{D}|} \sum_{j}^{|\mathcal{T}_i|}\mathds{1}[c^p_j \in \mathcal{G}^p_{k}] / |\mathcal{D}|.
\label{eq:proportional_grouping}
\end{equation}
The detailed algorithm for the proportional query grouping is provided in Alg. 1 of the supplementary material.
An example of the sum of predicate frequencies of a predicate group $\mathcal{G}^p_k$ and $|\mathcal{G}^q_k|$, the number of queries in each query group is in Tab.~\ref{tab:group_statistics}.
Proportional query grouping enables an output distribution to better resemble the GT distribution by design, where related results are presented in Fig.~\ref{fig:per_group_fig}.
\begin{table}[t]
    \centering
    \renewcommand{\arraystretch}{0.9}
     \setlength{\tabcolsep}{7pt}
    \vspace{-0.2cm}
    \begin{adjustbox}{max width=\linewidth}
    \begin{tabular}{c|cc}
    \toprule
    \small Group & \small \textbf{Number of GTs} & \small \textbf{Number of queries} \\
    \midrule
     \small Group 1  &  \small 990k (48.4\%)&  \small 146 (48.7\%)\\
     \small Group 2  &  \small 398k (19.5\%)&  \small 58 (19.3\%)\\
     \small Group 3  &  \small 299k (14.6\%)&  \small 43 (14.3\%)\\
     \small Group 4  &  \small 173k (8.5\%)&  \small 25 (8.3\%)\\
     \small Group 5  &  \small 186k (9.1\%)&  \small 28 (9.3\%)\\
     \bottomrule
    \end{tabular}
    \end{adjustbox}
    \vspace{-0.2cm}
    \caption{\textbf{Statistics of query groups when $N_g$ = 5.}}
    \label{tab:group_statistics}
    \vspace{-0.65cm}
\end{table}

\noindent\textbf{Groupwise query specialization.}
Given $N_g$ query groups $\{\mathcal{G}^q_i\}_{i=1}^{N_g}$ and predicate groups $\{\mathcal{G}^p_i\}_{i=1}^{N_g}$ defined as above, groupwise query specialization forces a query to only detect predicates in the corresponding predicate group.
In other words, a query in an $i$'th query group $\mathcal{G}^q_i$ is forced only to detect predicates in the $i$'th predicate group $\mathcal{G}^p_i$, instead of struggling to detect predicates in $\mathcal{G}^p_j$ where $i \neq j$.
To do so, we first define two mapping functions $\mathfrak{G}_p$ and $\mathfrak{G}_q$.
The first function $\mathfrak{G}_p$ returns the index of a predicate group in which a GT $t_i$ with a predicate label $c^p_i$ belongs to.
Similarly, the second function $\mathfrak{G}_q$ returns a query group in which a prediction $\hat{t}_j$ from a query $q_j$ belongs to as defined below:
\begin{equation}
\begin{split}
&\mathfrak{G}_p(t_i) = k\ \mbox{if}\ c^p_i \in \mathcal{G}^p_k, \\
&\mathfrak{G}_q(\hat{t}_j) = k\ \mbox{if}\ q_j \in \mathcal{G}^q_k.
\end{split}
\end{equation}
If a GT $t_i$ has a predicate label $c^p_i$ that belongs to the $k$'th predicate group, $\mathfrak{G}_p(t_i) = k$ holds. 
Similarly, if a predicted triplet $\hat{t}_j$ from a query $q_j$ that belongs to the $k$'th query group, $\mathfrak{G}_q(\hat{t}_j) = k$ holds.
Based on the mapping functions defined above, a grouping cost $\mathcal{H}_{\text{group}}$ is defined as:
\begin{equation}
    \mathcal{H}_{\text{group}}\left(t_i, \hat{t}_j\right) = 
    \begin{cases}
        0, &\mbox{if }\mathfrak{G}_p\left(t_i\right) = \mathfrak{G}_q\left(\hat{t}_j\right)\ \text{or} \ t_i= \varnothing, \\
        \infty, &\text{otherwise}.
    \end{cases}
\end{equation}
Then, the groupwise query specialization is done by adding the grouping cost $\mathcal{H}_{\text{group}}$ to the original matching cost $\mathcal{H}_{\text{match}}$ as below:
\begin{equation}
    \sigma^*_{\text{spec}} = \underset{\sigma \in \mathfrak{S}_{N_q}}{\mathrm{argmin}}\sum_i^{N_q}\mathcal{H}_{\text{match}}\left(t_i, \hat{t}_{\sigma(i)} \right) +     \mathcal{H}_{\text{group}}\left(t_i, \hat{t}_{\sigma(i)}\right) .
\label{eq:grouping}
\end{equation}
Adding $\mathcal{H}_{\text{group}}$ to the original matching cost results in a GT exclusively being assigned to predictions from the query with a group index identical to that of the GT, since the matching cost is set to $\infty$ otherwise.
\subsection{Quality-Aware Multi-Assignment}
\label{sec:sec_qg_dla}
The conventional assignment has a constraint that a GT can only be assigned to a single prediction. 
Due to the constraint, near-correct or even correct predictions are suppressed by being assigned a \q{no relation ($\varnothing$)} as a GT, which hinders proper training.
Therefore, we adopt the multi-assignment, which assigns a GT $t_i$ to $d_i$ number of high-quality predictions instead of only assigning it to a single prediction.

{\renewcommand{\arraystretch}{1.05}%
\begin{table*}[ht!]
    \centering
    \begin{adjustbox}{width=\textwidth}
    \begin{tabular}{l|cc||cc}
        \toprule
         Method & R@50/100 & mR@50/100 & AvgR@50/100 & F@50/100 \\
         \midrule
         \hline
         \rowcolor{black!15}
         \multicolumn{5}{l}{\textbf{\textit{X101-FPN backbone}}}\\
         Motifs \cite{zellers2018neural} &  32.1 / 36.9 & 5.5 / 6.8 &   18.8 / 21.9 & 9.4 / 11.5 \\
         VCTree \cite{tang2019learning} &  31.8 / 36.1 &   6.6 / 7.7 &  19.2 / 21.9 & 10.9 / 12.7 \\
         VCTree-TDE \cite{tang2020unbiased}   & 19.4 / 23.2 & 9.3 / 11.1 &  14.4 / 17.2 & 12.6 / 15.0 \\
         VCTree-EBM \cite{suhail2021energy}    & 20.5 / 24.7 &  9.7 / 11.6 & 15.1 / 18.2 & 13.2 / 15.8 \\
         VCTree-BPLSA \cite{guo2021general}   & 21.7 / 25.5 & 13.5 / 15.7 & 17.6 / 20.6 & 16.6 / 19.4 \\
         DT2-ACBS \cite{desai2021learning}   & 22.0 / 24.4 & 15.0 / 16.3 & 18.5 / 20.4  & 17.8 / 19.5 \\
         \hline
        \rowcolor{black!15}
         \multicolumn{5}{l}{\textbf{\textit{ResNet-101 backbone}}}\\
         RelDN \cite{zhang2019graphical,li2022sgtr}   &  30.3 / 34.8 & 4.4 / 5.4  &   17.4 / 20.1 & 7.7 / 9.3 \\
         BGNN \cite{li2021bipartite,li2022sgtr}   & 28.2 / 33.8 &  8.6 / 10.3 & 18.4 / 22.1  & 13.2 / 15.8 \\
         AS-Net \cite{chen2021reformulating}   &  18.7 / 21.1 & 6.1 / 7.2  &  12.4 / 14.2 & 9.2 / 10.7 \\
         SGTR \cite{li2022sgtr}              & 25.1 / 26.6 & 12.0 / 14.6 & 18.6 / 20.6 & 16.2 / 18.9 \\
         \midrule
         HOTR* \cite{kim2021hotr}   & 22.4 / 27.1 &  6.9 / 9.7  & 14.7 / 18.4 & 10.6 / 14.3  \\
         HOTR* + \textit{\textbf{SpeaQ (Ours)}}   & 24.7{\small \textcolor{red}{(+2.3)}} / 29.1{\small \textcolor{red}{(+2.0)}} & 9.6{\small \textcolor{red}{(+2.7)}} / 12.7{\small \textcolor{red}{(+3.0)}} & 17.2{\small \textcolor{red}{(+2.5)}} / 20.9{\small \textcolor{red}{(+2.5)}} & 13.8{\small \textcolor{red}{(+3.2)}} / 17.7{\small \textcolor{red}{(+3.4)}} \\
         \midrule
         ISG$^*_\dagger$~\cite{khandelwal2022iterative}      &  29.5 / 32.1 & 7.4 / 8.4 & 18.5 / 20.3 & 11.8 / 13.3 \\
         ISG$^*_\dagger$+ \textbf{\textit{SpeaQ (Ours)}} & 
         \textbf{32.9}{\small \textcolor{red}{(+3.4)}} / 
         \textbf{36.0}{\small \textcolor{red}{(+3.9)}} & 
         11.8{\small \textcolor{red}{(+4.4)}} /
         14.1{\small \textcolor{red}{(+5.7)}} &
         22.4{\small \textcolor{red}{(+3.9)}} /
         25.1{\small \textcolor{red}{(+4.8)}} & 
         17.4{\small \textcolor{red}{(+5.6)}} /
         20.3{\small \textcolor{red}{(+7.0)}} \\
         \midrule
         ISG$^*$      &  27.2 / 30.1 & 15.0 /  16.6 & 21.1 / 23.4  & 19.3 / 21.4 \\
         ISG$^*$+ \textbf{\textit{SpeaQ (Ours)}} & 
         32.1{\small \textcolor{red}{(+4.9)}} / 
         35.5{\small \textcolor{red}{(+5.4)}}& 
         \textbf{15.1}{\small \textcolor{red}{(+0.1)}} /
         \textbf{17.6}{\small \textcolor{red}{(+1.0)}} & 
         \textbf{23.6}{\small \textcolor{red}{(+2.5)}} / 
         \textbf{26.6}{\small \textcolor{red}{(+3.2)}}  & 
         \textbf{20.5}{\small \textcolor{red}{(+1.2)}} / 
         \textbf{23.5}{\small \textcolor{red}{(+2.1)}} \\
        \bottomrule
    \end{tabular}
    \end{adjustbox}
        \caption{\textbf{Performance on Visual Genome.}
    The best results among models with ResNet-101 backbone are marked in \textbf{bold}. * denotes reproduced results. $\dagger$ denotes the performance without loss re-weighting proposed in~\cite{khandelwal2022iterative}.}
    \label{tab:main_table}
    \vspace{-0.5cm}
\end{table*}
}
\noindent\textbf{Triplet quality-aware determination of $d_i$.}
Since the number of high-quality predictions that correspond to a GT $t_i$ may vary, we determine $d_i$ adaptively considering the overall triplet-level prediction quality of a subject, object, and the predicate on a GT $t_i$, instead of setting $d_i$ equal for every GT.
In detail, vectors that represent subject, object and predicate prediction qualities between a GT $t_i$ and every prediction $\{\hat{t}_j\}_{j=1}^{N_q}$ are firstly calculated as:
\begin{equation}
\begin{split}
&v^s_i=\left[\text{IoU}\left(b^s_{i}, \hat b^s_{j}\right)\right]_{j=1}^{N_q} \in \mathbb{R}^{N_q}, \\
&v^o_i=\left[\text{IoU}\left(b^o_{i}, \hat b^o_{j}\right)\right]_{j=1}^{N_q} \in \mathbb{R}^{N_q}, \\
&v^r_i = \left[\hat{c}^p_j(c^p_i)\right]^{N_q}_{j=1} \in \mathbb{R}^{N_q}, \\
\end{split}
\label{eq:k_kernel}
\end{equation}
where $\text{IoU}$ is a function that outputs an IoU between two bounding boxes, and $\hat{c}^p_j(c^p_i)$ denotes the predicted probability of a GT predicate label $c^p_i$ of a prediction $\hat{t}_j$.
The $j$'th element in the resulting vectors represents the prediction quality of $\hat{t}_j$ on a GT $t_i$.
Concretely, for a GT triplet $t_i$, IoU between the GT subject box $b^s_i$ and predicted subject boxes $\hat{b}^s_j$ from every prediction $\hat{t}_j$ is calculated to form a subject quality vector $v^s_i \in \mathbb{R}^{N_q}$, where object quality vector $v^o_i \in \mathbb{R}^{N_q}$ is also analogously defined.
Moreover, a predicate quality vector $v_i^r$ is defined as a predicted score of a GT predicate label.
Then, $d_i$ is calculated given subject, object and predicate quality vectors as below:
\begin{equation}
\begin{split}
&v_i=\mathcal{R}\left(v^s_i, v^o_i\right) + \lambda_{\text{rel}} v_i^r \in \mathbb{R}^{N_q},\\
&d_i=\left\lfloor\max\left(\sum\text{top-k}\left(v_i\right),1\right)\right\rfloor \in \mathbb{N},
\label{eq:d_i_calculate}
\end{split}
\end{equation}
where $\mathcal{R}$ is the element-wise function (\eg $\text{min, max}$), and top-k is a function that only retains $k$ largest elements in the vector, and sets the value as zero otherwise.
The triplet-level quality vector $v_i$ is firstly obtained by combining the output of a relation function $\mathcal{R}$ and the predicate quality vector $v_i^r$, and then fed into the top-k function.
Then, the floored result of the sum of elements in the resulting vector from the top-k function is set as $d_i$ if the sum is larger than 1.
Otherwise, $d_i$ is set as 1 to ensure that every GT is assigned to a prediction at least once.

\noindent\textbf{Quality-aware multi-assignment.}
With $d_i$ calculated in a triplet quality-aware manner as elucidated above, an augmented GT set $\mathcal{T}_{i'}$ is constructed by duplicating $t_i$ for $d_i$ times and padding $\varnothing$ until $|\mathcal{T}_{i'}|$ reaches $N_q$.
Then, the quality-aware multi-assignment is formally defined as:
\begin{equation}
\sigma^*_{\text{multi}} = \underset{\sigma \in \mathfrak{S}_{N_q}}{\mathrm{argmin}}\sum_{i'}^{N_q}\mathcal{H}_{\text{match}}\left(t_{i'}, \hat{t}_{\sigma(i')} \right).
\label{eq:o2m}
\end{equation}
Given the objective above, Hungarian algorithm finds the permutation $\sigma^*_{\text{multi}}$ with the lowest matching cost between the augmented GT set $\mathcal{T}_{i'}$ and the prediction set $\hat{\mathcal{T}}_{i}$. 

\noindent\textbf{Final training objective.}
Proposed groupwise query specialization and quality-aware multi-assignment are combined to form the final assignment objective, dubbed as Groupwise Query \textbf{Spe}ci\textbf{a}lization and \textbf{Q}uality-Aware Multi-Assignment (SpeaQ) as follows:
\begin{equation}
\sigma^* = \underset{\sigma \in \mathfrak{S}_{N_q}}{\mathrm{argmin}}\sum_{i'}^{N_q}\mathcal{H}_{\text{match}}\left(t_{i'}, \hat{t}_{\sigma(i')} \right) +     \mathcal{H}_{\text{group}}\left(t_{i'}, \hat{t}_{\sigma(i')}\right) .
\label{eq:total}
\end{equation}
With the final matching cost defined above, the optimal permutation of predictions $\sigma^*$ with the lowest matching cost is obtained.
Then, the final training loss is defined as the sum of the subject, predicate and object loss, $\mathcal{L}_{\text{total}}\left(t_{i'}, \hat{t}_{\sigma^*(i')} \right) = \mathcal{L}_{\text{s}}\left(t_{i'}, \hat{t}_{\sigma^*(i')} \right) + \mathcal{L}_{\text{p}}\left(t_{i'}, \hat{t}_{\sigma^*(i')} \right) + \mathcal{L}_{\text{o}}\left(t_{i'}, \hat{t}_{\sigma^*(i')} \right)$.
The subject loss $\mathcal{L}_{\text{s}}\left(t_{i'}, \hat{t}_{\sigma^*(i')} \right)$ is defined as:
\begin{equation}
\begin{split}
    & \mathcal{L}_{\text{s}}\left(t_{i'}, \hat{t}_{\sigma^*(i')} \right) \\
    & = \mathcal{L}_{cls}(c^s_{i'}, \hat{c}^s_{\sigma^*(i')}) + \mathds{1}_{\left\{t_{i'} \neq \varnothing \right\}}\mathcal{L}_{box}(b^s_{i'}, \hat{b}^s_{\sigma^*(i')}),
\end{split}
\end{equation}
where $\mathcal{L}_{cls}(c^s_{i'}, \hat{c}^s_{\sigma^*(i')})$ is a classification loss (\ie cross-entropy) between the subject label and the predicted subject logit and $\mathcal{L}_{box}(b^s_{i'}, \hat{b}^s_{\sigma^*(i')})$ is a sum of regression losses (\ie L1 loss and GIoU loss) between the ground-truth bounding box and the predicted bounding box.
$\mathcal{L}_{\text{p}}\left(t_{i'}, \hat{t}_{\sigma^*(i')} \right)$ and $\mathcal{L}_{\text{o}}\left(t_{i'}, \hat{t}_{\sigma^*(i')} \right)$ are analogously defined.

\section{Experiments}
In this section, we compare the performance of the proposed SpeaQ with state-of-the-art methods for Scene Graph Generation and Human-Object Interaction Detection tasks.
Further implementation details are in the Sec. B of the supplementary material.
\subsection{Datasets}
\noindent\textbf{Visual Genome.}
Visual Genome dataset consists of 108k images with 75k objects and 37k predicates. 
Following previous works~\cite{xu2017scene,zellers2018neural}, we use the subset of Visual Genome (\ie VG150), which is composed of the most frequent 150 objects and 50 predicate categories.
We report the performance on two widely adopted metrics Recall@K (R@K) and Mean Recall@K (mR@K).
Since R@K and mR@K are known to be biased toward the most frequent and the least frequent classes, we also report the arithmetic mean (AvgR@K)~\cite{li2022devil} and the harmonic mean (F@K)~\cite{khandelwal2022iterative} of two metrics to measure the overall balanced performance across predicate frequency.
\begin{table}[t]
    \centering
    \resizebox{\columnwidth}{!}{
    \setlength{\tabcolsep}{3.5pt}  
    \begin{tabular}{l  c  c c }
         \toprule
         Method & 
         $\  $ {Full} $\  $ & 
         {Rare} & 
         {Non-Rare}
         \\
         \midrule
         {UnionDet~\cite{Kim2020UnionDet}}  &14.25 & 10.23 & 15.46 \\
         {PastaNet~\cite{li2020pastanet}} &22.65 & 21.17 & 23.09  \\
         {IDN~\cite{yonglu2020idn}}  & 23.36 & 22.47 & 23.63 \\
         {HOITrans~\cite{zou2021end}}   & 23.46 & 16.91 & 25.41 \\
        {HOTR~\cite{kim2021hotr}}  & 25.10 & 17.34 & 27.42 \\
        {AS-Net~\cite{chen2021reformulating}}   & 28.87 & 24.25 & 30.25 \\
        {QPIC~\cite{tamura2021qpic}}  & 29.07 & 21.85 & 31.23 \\
        {MSTR~\cite{kim2022mstr}}  & 31.17 & 25.31 & 32.92 \\
        {CDN~\cite{zhang2021mining}} & 31.44 & 27.39 & 32.64 \\
        {UPT~\cite{zhang2022efficient}}  & 31.66 & 25.94 & 33.36 \\
         \midrule
        GEN$^*$~\cite{liao2022gen} &  33.12 & 27.12 & 34.91 \\ 
        GEN$^*$ + \textbf{\textit{SpeaQ (Ours)}} & \textbf{34.00}{\scriptsize \textcolor{red}{(+0.88)}}&\textbf{30.20}{\scriptsize \textcolor{red}{(+3.08)}}&\textbf{35.13}{\scriptsize \textcolor{red}{(+0.22)}}  \\
         
         \bottomrule
    \end{tabular}
    }
    \vspace{-0.2cm}
    \caption{\textbf{Performance on HICO-DET.} Best results are marked in \textbf{bold}. * denotes reproduced result.}
    \vspace{-0.3cm}
    \label{table:hico_table}
\end{table}

\begin{table}[t!]
    \centering
    \renewcommand{\arraystretch}{0.85}
    \setlength{\tabcolsep}{8pt}
    \begin{adjustbox}{max width=\linewidth}
    \begin{tabular}{ll|cc||c}
         \toprule
           G &   Q &   R@100 &   mR@100 &   AvgR@100\\
         \midrule
            &    &   32.1 &   8.4 &   20.3\\
            &   \cmark &   33.1 &   9.2 &   21.2\\
           \cmark &    &   35.3 &   13.5 &   24.4\\
           \cmark &   \cmark &   \textbf{36.0} &   \textbf{14.1} &   \textbf{25.1}\\
         \bottomrule
    \end{tabular}
    \end{adjustbox}
    \vspace{-0.1cm}
    \caption{  \textbf{Ablation study on main components.} G : Groupwise Query Specialization (Sec.~\ref{sec:sec_query_grouping}), Q : Quality-Aware Multi-Assignment (Sec.~\ref{sec:sec_qg_dla})}
    \vspace{-0.6cm}
    \label{tab:main_ablation}
\end{table}

\noindent\textbf{HICO-DET.}
For the Human-Object Interaction Detection task, we report the performance on the HICO-DET~\cite{chao2018learning} benchmark, which contains 47k images (37.5k for training and 9.5k for testing) with more than 150k annotated instances of human-object pairs. 
It has 600 triplet classes, a subset of possible combinations between 80 instance classes and 117 verb classes.
We report mAP on three different sets: \textit{full} including all 600 classes, \textit{rare} including 138 classes having less than 10 training instances and \textit{non-rare} including 462 classes with more than 10 training samples.
\subsection{Experimental Results}
\paragraph{Results on Visual Genome.}
In Tab.~\ref{tab:main_table}, we report the performance of SpeaQ when applied to two competitive Transformer-based models, ISG~\cite{khandelwal2022iterative} and HOTR~\cite{kim2021hotr}.
Applying SpeaQ on ISG results in a gain of 5.4 and 1.0 on R@100 and mR@100, achieving the state-of-the-art result.
The result is remarkable in that no previous work has achieved the best performances on both of the two contradicting metrics R@100 and mR@100, which is shown by the best results on AvgR@100 and F@100.
Similarly, when applied to HOTR, improvements of 2.0 and 3.0 on R@100 and mR@100 are reported.
Consistent gains in both models show the generalizability of the proposed SpeaQ.
Note that the boost in performance is gained with \textit{zero} additional model parameters or inference cost.

\noindent\textbf{Results on HICO-DET.} We also validate the effectiveness and generalizability of SpeaQ by applying SpeaQ on top of a competitive baseline in the Human-Object Interaction (HOI) Detection task, GEN~\cite{liao2022gen}.
The result is reported in Tab.~\ref{table:hico_table}.
Applying SpeaQ to GEN results in a gain of 0.88, 3.08, and 0.22 in the full, rare, and non-rare sets, respectively.
Again, the result shows that applying SpeaQ results in a consistent gain in performance across various tasks and models.

\noindent\textbf{Ablation study on main components.}
In Tab.~\ref{tab:main_ablation}, ablation results of the proposed components are reported.
Quality-aware multi-assignment consistently boosts both performances on R@100 and mR@100 by 1.0 and 0.8 when applied to the baseline.
Also, groupwise query specialization results in a gain of 3.2 and 5.1 on R@100 and mR@100 when applied to the baseline.
With both components combined, the best performance with 36.0 of R@100 and 14.1 of mR@100 is attained.
\section{Analysis}
\label{sec:analysis}
\begin{figure}[t!]
\centering

\begin{minipage}{0.48\linewidth}
    \centering
    \includegraphics[width=\linewidth]{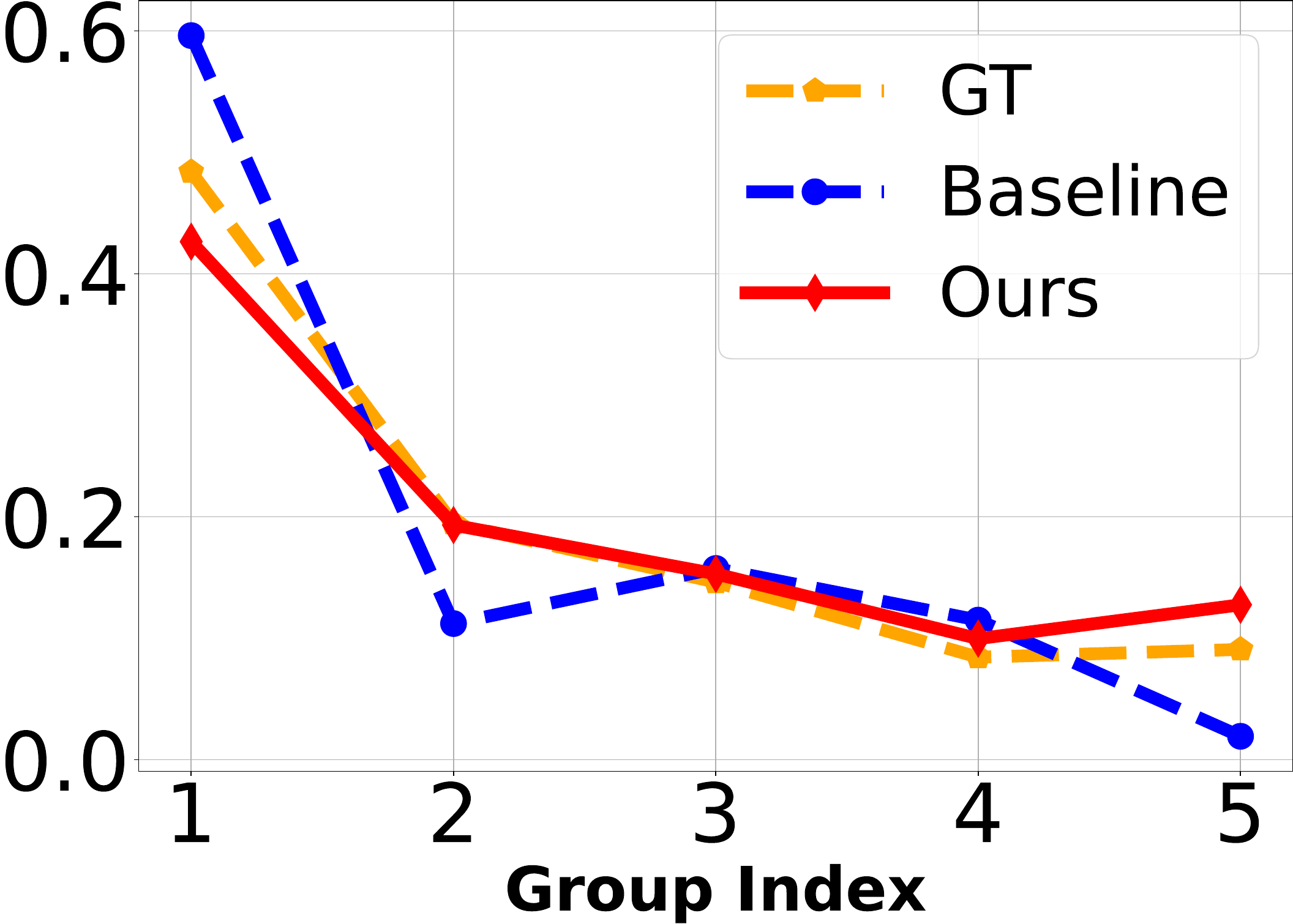}
    \subcaption{\scriptsize \textbf{Prediction frequency} per group. The closer to the GT, the better.}
    \label{fig:pred_distribution_per_group_fig}
\end{minipage}
\hfill
\begin{minipage}{0.48\linewidth}
    \centering
    \includegraphics[width=\linewidth]{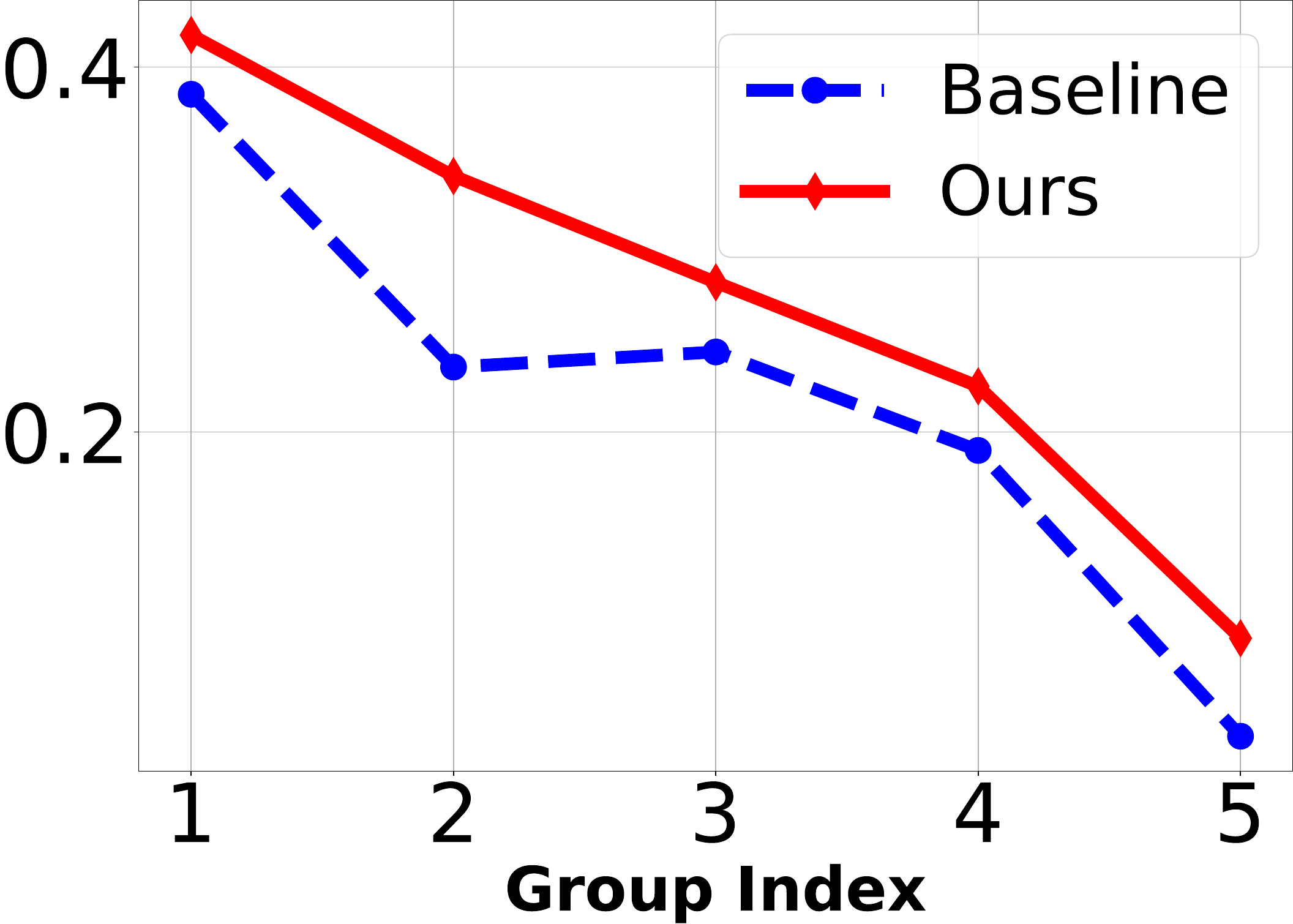}
    \subcaption{\scriptsize \textbf{mR@100} per group. The higher, the better.}
    \label{fig:mr_100_per_group_fig}
\end{minipage}
\vspace{-0.2cm}
\caption{\small \textbf{Prediction frequency and mR@100} per group. Group 1 consists of the most frequent predicates, while group 5 consists of the least frequent predicates.}
\vspace{-0.6cm}
\label{fig:per_group_fig}

\end{figure}

\begin{table*}[ht!]
\centering
\renewcommand{\arraystretch}{0.9}
\setlength{\tabcolsep}{7pt}
\begin{tabular}{l|cccccc|ccccccc}
\toprule
\multirow{2}{*}{Metrics} & \multicolumn{6}{c|}{$N_g$} & \multicolumn{7}{c}{$k$}\\
& 1 & 2 & 3 & 4 & 5 & 6 & 1 & 2 & 3 & 4 & 5 & 6 & 7\\
\midrule
mR@100 & 8.8 & 12.2 & 12.6 & 14.1 & 14.2 & \textbf{14.3} & 8.8 & 12.4 & 12.9 & 13.5 & 13.8 & \textbf{14.1} & 13.8 \\
R@100 & 33.0 & 35.6 & 35.6 & \textbf{36.0} & 35.7 & 35.7 & 33.0 & 35.9 & 35.8 & 36.0 & \textbf{36.1} & 36.0 & 35.9 \\
\hline
\midrule
AvgR@100 & 20.9 & 23.9 & 24.1 & \textbf{25.1} & 25.0 & 25.0 & 20.9 & 24.2 & 24.4 & 24.8 & 25.0 & \textbf{25.1} & 24.9 \\

\bottomrule
\end{tabular}
\vspace{-0.2cm}
\caption{\textbf{Performance under different numbers of groups $N_g$, and $k$ for top-k function in Eq.~\eqref{eq:d_i_calculate}.} 
}
\label{tab:n_g_n_k_ablation}
\vspace{-0.3cm}
\end{table*}
\begin{table}[t]
    \centering
     \renewcommand{\arraystretch}{0.95}
     \setlength{\tabcolsep}{7pt}
    \begin{adjustbox}{max width=\linewidth}
    \begin{tabular}{c|c|cc||c}
         \toprule
            Method &   $N_q$ &    R@100 &    mR@100 &  
            AvgR@100 \\
         \midrule
            Baseline &   300 &  32.1 &  8.4 &  20.3\\
            Baseline &   600 &  32.3\textcolor{red}{(+0.2)} &  7.8\textcolor{blue}{(-0.6)} &  20.1\textcolor{blue}{(-0.2)}\\
         \midrule
            Ours &  300 &  35.8 &  13.0 &  24.4 \\
            Ours &  600 &  \textbf{36.0}\textcolor{red}{(+0.2)} &   \textbf{14.1}\textcolor{red}{(+1.1)} &  \textbf{25.1}\textcolor{red}{(+0.7)} \\
         \bottomrule
    \end{tabular}
    \end{adjustbox}
    \vspace{-0.3cm}
    \caption{  \textbf{Performance with a larger number of queries $N_q$.}}
    \label{tab:query_num_ablation}
    \vspace{-0.3cm}
\end{table}
In this section, we present various experimental results along with analyses to validate the effectiveness of the SpeaQ.
\q{Baseline} in all experiments denotes ISG~\cite{khandelwal2022iterative}.
Note that all experiments are done without the loss re-weighting proposed in~\cite{khandelwal2022iterative} to focus on the effect of components since it largely biases a model toward rare predicates.

\noindent\textbf{Analyses on output frequency and mR@100 per group.}
In Fig.~\ref{fig:per_group_fig}, the output frequency and mR@100 per predicate group are plotted.
Note that we define mR@100 of a group as an average of recall of predicates in a group, similar to the definition of conventional mR@100.
By applying SpeaQ, an overprediction of frequent classes and an underprediction of rare classes are relieved as shown in Fig.~\ref{fig:pred_distribution_per_group_fig}, resulting in an output distribution (\textcolor{red}{red}) closer to the GT distribution (\textcolor{orange}{orange}) in every group compared to the baseline, which was initially biased toward frequent predicates (\textcolor{blue}{blue}).
The effect is shown in Fig.~\ref{fig:mr_100_per_group_fig}, where consistent performance gains in every group are reported.
It is notable that the performance improves in the most frequent group while the prediction frequency declines, which shows that specialized queries are better at performing a task even with a smaller number of predictions compared to unspecialized queries.
Overall, results show that the specialization of queries improves the performance of own target task of a query, and the collection of specialized queries better resembles the GT distribution.

\noindent\textbf{Applying SpeaQ to the model with a larger $N_q$.}
In Tab.~\ref{tab:query_num_ablation}, performances of the baseline and the model trained with SpeaQ under a different number of queries $N_q$ are reported.
Naively enlarging $N_q$ under a conventional training scheme results in a drop of 0.6 on mR@100, which is three times larger than the gain of 0.2 on R@100.
In contrast, both R@100 and mR@100 are improved by 0.2 and 1.1 as enlarging $N_q$ when trained with SpeaQ. 
The result validates that SpeaQ is better at fully leveraging a model's capacity by training specialized queries compared to the baseline which fails to successfully handle the model capacity, therefore benefits by scaling up the number of queries.

\begin{table}
    \centering
     \renewcommand{\arraystretch}{0.95}
     \setlength{\tabcolsep}{7pt}
    \begin{adjustbox}{max width=0.9\linewidth}
    \begin{tabular}{c|cc||c}
    \toprule
        Method &  R@100 &  mR@100 &  AvgR@100 \\
    \midrule
        Baseline & 32.1 & 8.4 & 20.3 \\
    \midrule
        Uniform &  33.8 &  13.2 &  23.5 \\
        Proportional & \textbf{36.0} &  \textbf{14.1} &  \textbf{25.1}\\
    \bottomrule
  \end{tabular}
  \end{adjustbox}
\vspace{-0.3cm}
\caption{  \textbf{Ablation study on proportional query grouping.}}
\label{tab:proportional_grouping_tab}
\vspace{-0.5cm}
\end{table}
\noindent\textbf{Analysis on $N_g$.}
In Tab.~\ref{tab:n_g_n_k_ablation}, experimental results under different numbers of groups $N_g$ are reported.
Compared to the baseline ($N_g$ = 1), the performance gain is reported regardless of $N_g$.
AvgR@100 gradually improves as $N_g$ enlarges, and plateaus at $N_g$ = 4 then slightly decreases afterward.
Based on the result, we suppose that partitioning queries into an overly large number of groups may be suboptimal since the lack in the amount of GTs in a group may result in insufficient training signals.

\begin{figure*}[t!]
\centering
\includegraphics[width=\textwidth]{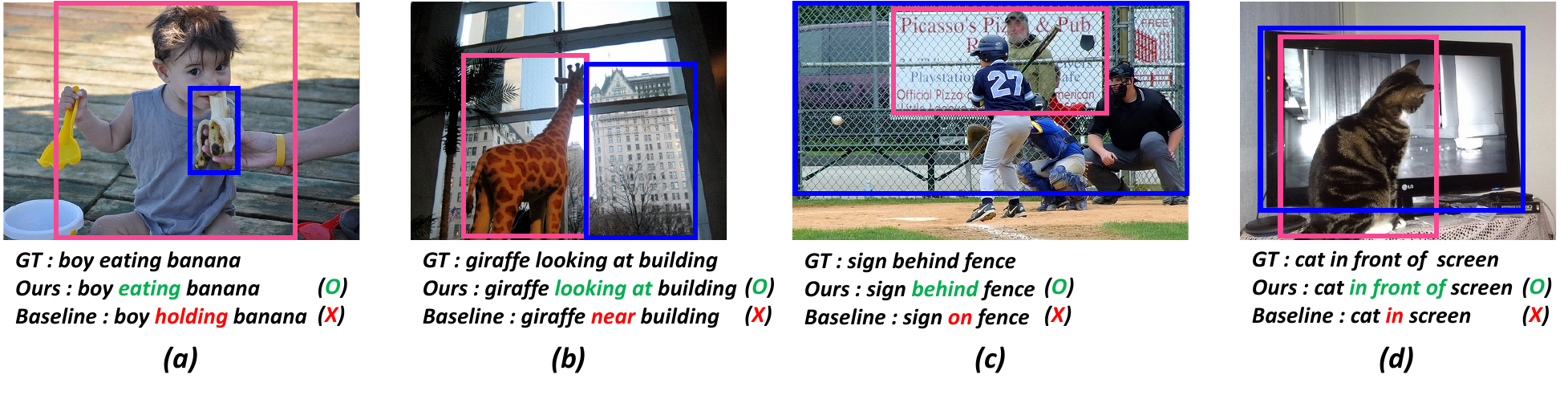}
\vspace{-0.75cm}
\caption{
\textbf{Qualitative results on Visual Genome dataset.} Predictions of the baseline and the model trained with SpeaQ are visualized along with corresponding ground-truths. Correct and wrong prediction results are marked {\color{ForestGreen}{green}} and \textcolor{red}{red}, respectively.}
\label{fig:qual_figure}
\vspace{-0.5cm}
\end{figure*}

\noindent\textbf{Analysis on $k$.}
In Tab.~\ref{tab:n_g_n_k_ablation}, experimental results under different $k$ for a top-k function used to calculate $d_i$ in Eq.~\eqref{eq:d_i_calculate} are reported.
The result shows that providing further positive signals consistently boosts the performance compared to the baseline ($k=1$), robust to the choice of $k$. 
The best performance is achieved when $k=6$ and the performance slightly decreases afterward, since overly large $k$ may provide positive signals even to non-promising predictions.

\begin{table}
\centering
\renewcommand{\arraystretch}{0.8}
\setlength{\tabcolsep}{7pt}
    \begin{tabular}{c|c|cc||c}
    \toprule
       \small Criterion & \small F & \small R@100 &  \small mR@100 & \small AvgR@100 \\
    \midrule
       \small Baseline & \xmark &\small 32.1 & \small  8.4 &\small  20.3 \\
    \midrule
        \small Random & \xmark &\small 30.7 &\small 11.4 &\small 21.1 \\
        \small Semantic & \xmark &\small 31.8 &\small 10.0 &\small 20.9 \\
        \small BGNN~\cite{li2021bipartite} & \cmark &\small 32.7 &\small 12.7 &\small 22.7 \\
        \small SHA~\cite{dong2022stacked} & \cmark &\small 34.1 &\small  13.4 &\small 23.8 \\
     \midrule
        \small Ours & \cmark & \small \textbf{36.0} & \small \textbf{14.1} & \small \textbf{25.1}\\
    \bottomrule
  \end{tabular}
  \vspace{-0.3cm}
 \caption{  \textbf{Analysis on predicate grouping criterion.} \q{F} denotes that the predicates are grouped on a frequency-basis.}
 \vspace{-0.4cm}
 \label{tab:grouping_criterion}
\end{table}
\begin{table}
    \centering
     \renewcommand{\arraystretch}{0.9}
     \setlength{\tabcolsep}{7pt}
    \begin{adjustbox}{max width=0.9\linewidth}
    \begin{tabular}{c|cc||c}
    \toprule
       $\mathcal{R}$ &  R@100 &  mR@100 &  AvgR@100 \\
    \midrule
        Baseline & 32.1 & 8.4 & 20.3 \\
    \midrule
        min &  35.6 &  12.7 &  24.2\\
        mean & \textbf{36.0} &  13.3 &  24.7\\
        max &  \textbf{36.0} &  \textbf{14.1} &  \textbf{25.1}\\
    \bottomrule
  \end{tabular}
  \end{adjustbox}
\vspace{-0.3cm}
\caption{  \textbf{Analysis on relation function $\mathcal{R}$.}}
\label{tab:ablation_relation_function}
\vspace{-0.7cm}
\end{table}
\noindent\textbf{Ablation study on proportional query grouping.}
In Tab.~\ref{tab:proportional_grouping_tab}, the performances of the proportional query grouping (Eq.~\eqref{eq:proportional_grouping}) compared to the uniform query grouping are reported, where the uniform query grouping denotes an equal number of queries assigned to every query group.
Results show that while a uniform grouping improves the performance compared to the baseline by training specialized queries, the best performance is achieved under the proportional query grouping.
Based on the result, we suggest that providing a balanced amount of supervision to every query on average helps better train queries.

\noindent\textbf{Analysis on predicate grouping criterion.}
In Tab.~\ref{tab:grouping_criterion}, results under different predicate grouping criteria are reported.
\q{Random} denotes predicates are randomly grouped into five groups with equal size, and \q{semantic} denotes predicates are divided into three groups (\q{geometric}, \q{possessive}, and \q{semantic}) by their lexical semantics~\cite{zellers2018neural}.
Also, we report performance under adopting predicate groups from previous works~\cite{li2021bipartite,dong2022stacked} split on a frequency-basis.
Further details about predicate groups are in Sec.~C in the supplementary material.
Results show that adopting the frequency-basis group as $\mathcal{G}^p_i$ consistently outperforms random or semantic criterion regardless of the choice of the grouping strategy, since it relieves training difficulties caused by imbalanced training signals, while the proposed grouping criterion results in the best performance.

\noindent\textbf{Analysis on choice of $\mathcal{R}$.}
In Tab.~\ref{tab:ablation_relation_function}, performances under adopting min, mean, and max as $\mathcal{R}$ in Eq.~\eqref{eq:d_i_calculate} are reported.
The best result is reported in both R@100 and mR@100 when adopting max as $\mathcal{R}$, followed by mean and min functions.
Based on the result, we suggest that providing a chance for cases where only a single instance is correctly detected helps the model better learn samples that the model is confused about (\ie max) than conservatively rewarding \q{perfectly} detected cases (\ie min).

\begin{table}[t]
\begin{minipage}{\linewidth}
    \centering
     \renewcommand{\arraystretch}{0.95}
     \setlength{\tabcolsep}{7pt}
    \resizebox{\textwidth}{!}{
    \begin{tabular}{c|c|cc||c}
    \toprule
       Type &   avg($d_i$)  &   R@100 &   mR@100 &   AvgR@100\\
    \midrule
        IoU &  23.4 &  3.2 &  5.0&  4.1\\
       Single &   1 &   32.1 &   8.4 &  20.3\\
       Agnostic &  3  &  33.0 &  9.0  &  21.0\\
     \midrule
       Ours &  3.2  &  \textbf{33.1} &   \textbf{9.2} &  \textbf{21.2}\\
    \bottomrule
  \end{tabular}
}
\vspace{-0.2cm}
\caption{ \textbf{Performances under various label assignment strategies.} avg($d_i$) : Average number of predictions a GT is assigned to.}
\label{tab:o2m_ablation_tab}
\vspace{-0.7cm}
\end{minipage}
\end{table}

\noindent\textbf{Quantitative results of quality-aware multi-assignment.}
Experimental results in Tab.~\ref{tab:o2m_ablation_tab} support the effectiveness of the quality-aware multi-assignment.
We compare performance when adopting conventional assignment, quality-agnostic multi-assignment, and quality-aware multi-assignment as a label assignment strategy, where each is denoted as single, agnostic, and ours.
A quality-agnostic multi-assignment denotes that $d_i$ in Eq.~\eqref{eq:d_i_calculate} is set equal to every GT.
We also report the performance under a simple IoU-based assignment commonly adopted in CNN-based detectors, where a GT is assigned to predictions with IoU over 0.5 for both subject and object.
The result shows that an IoU-based assignment completely fails in training Transformer-based models.
In contrast, quality-agnostic multi-assignment improves the performance compared to a single assignment, while quality-aware multi-assignment further improves the performance showing the best result on R@100 of 33.1 and mR@100 of 9.2.
The result shows the effectiveness of multi-assignment, and it could be further improved with quality-aware determination of $d_i$.
For better understanding, we further provide an intuitive running example that demonstrates the importance of the proposed assignment in Sec.~D of the supplementary material.

\noindent\textbf{Qualitative results.}
Fig.~\ref{fig:qual_figure} presents qualitative examples comparing prediction results from the baseline model and SpeaQ. 
Regarding samples (a) and (b) of the figure, the model trained with SpeaQ successfully detects challenging samples that require a detailed understanding of both the predicate's semantics and the image. 
This is in contrast to the baseline model, which struggles in these samples. 
Furthermore, as shown in samples (c) and (d), SpeaQ helps the model detect less common predicates. 
Concretely, the model trained with SpeaQ correctly classifies \q{behind} and \q{in front of}, which are 17 and 18 times less frequent compared to \q{on} and \q{in}, predicted by the baseline.
These improvements are attributed to queries trained with SpeaQ, which are specialized in target predicates therefore are better at detecting rare and challenging samples.
\section{Conclusion}
\vspace{-0.2cm}
In this paper, we propose a Groupwise Query \textbf{Spe}ci\textbf{a}lization and \textbf{Q}uality-Aware Multi-Assignment (SpeaQ).
The first component trains a \q{specialized} query by dividing queries and relations into groups and directing a query in a specific query group solely toward relations in the corresponding relation group.
The second component provides abundant training signals considering the triplet-level  quality of multiple predictions.
Our experiments show that SpeaQ results in performance gains across multiple VRD models and benchmarks with zero additional inference cost.

\noindent\textbf{Acknowledgements.}
This work was partly supported by ICT Creative Consilience Program through the Institute of Information \& Communications Technology Planning \& Evaluation(IITP) grant funded by the Korea government(MSIT)(IITP-2024-2020-0-01819), the National Research Foundation of Korea(NRF) grant funded by the Korea government(MSIT)(NRF-2023R1A2C2005373), the Electronics and Telecommunications Research Institute (ETRI) grant funded by the Korean government (24ZB1200, Fundamental Technology Research for Human-Centric Autonomous Intelligent Systems), and by Neubla.

{
    \small
    \bibliographystyle{ieeenat_fullname}
    \bibliography{main}
}
\clearpage
\appendix
\section*{\Large Appendix}
\section{Details of frequency-based predicate grouping and proportional query grouping.}
\label{para:query_grouping}
In this section, we provide a detailed algorithm for dividing $C^p = \{c^p_l\}^{N_p}_{l=1}$, a set of $N_p$ predicates, and $\mathcal{Q}$, a set of $N_q$ queries into $N_g$ groups, as introduced in Sec. 3.2 of the main paper.
First, $f_p(c^p_l)$, a proportion of a predicate $c^p_l$ is defined as the number of training samples having $c^p_l$ as a predicate label divided by the total number of training samples.
Similarly, the frequency of a predicate group $\mathcal{G}^p_i$ is then defined as the sum of frequencies of predicates in the group:
\begin{equation}
\label{eq:relative_freq}
    \begin{split}
        & f_g(\mathcal{G}^p_i) = \underset{c^p_l \in \mathcal{G}^p_i}{\sum} f_p(c^p_l).
    \end{split}
\end{equation}
Then, the algorithm iteratively assigns a predicate to a predicate group from the most frequent predicates to rare ones.
A predicate $c^p_l$ is assigned to the current predicate group $\mathcal{G}^p_i$ if the value of $f_g(\mathcal{G}^p_i)$ does not exceed a threshold of $(\frac{1}{2})^i$, where $i$ is initialized as 1. 
Otherwise, the predicate group index is incremented to move to the next group, and then the predicate is assigned to the new group.
This process continues until all predicates have been allocated or until the $N_{g}-1$'th group is filled. 
If there still exist remaining predicates, those are allocated to the final predicate group $\mathcal{G}^p_{N_g}$.
After predicate groups are formed, the number of queries in each group $|\mathcal{G}^q_k|$ is determined as a floored result of $f_g(\mathcal{G}^p_k)$ multiplied by the number of queries $N_q$, where $k = 1, ..., N_g$.
Similarly, the remaining queries after the allocation is done are assigned to the last query group $\mathcal{G}^q_{N_g}$.
The pseudocode is presented in Algorithm~\ref{alg:predicate_div}.
\begin{algorithm*}[ht]
\caption{Frequency-based Predicate Grouping and Proportional Query Grouping}\label{alg:predicate_div}
\begin{algorithmic}[1]
\State Initialize a predicate group index $i = 1$ 
\State Create an empty predicate group $\mathcal{G}^p_i$
\State Sort the predicate classes by frequency in descending order using the training set
\While{$i < N_g$}
\For{predicate $c^p_l$ in the sorted predicate list} \Comment{Frequency-based Predicate Grouping}
    \If{$f_g\left(\mathcal{G}^p_i\right) \leq (\frac{1}{2})^i$}
        \State Add $c^p_l$ to $\mathcal{G}^p_i$
    \Else
            \State Increment $i$ by $1$
            \State Create an empty predicate group $\mathcal{G}^p_i$
            \State Add $c^p_l$ to $\mathcal{G}^p_i$
    \EndIf
\EndFor
\EndWhile
\If{there are predicates remaining}
    \State Add remaining predicates to a predicate group $\mathcal{G}^p_{N_g}$
\EndIf

\For{query group index $k$ in $k = 1, 2, ... N_g$} \Comment{Proportional Query Grouping}
    \State Create an empty query group $\mathcal{G}^q_k$
    \State Assign $\lfloor N_q f_g(\mathcal{G}^p_k)\rfloor$ queries to a query group $\mathcal{G}^q_k$
\EndFor
\If{there are queries remaining}
    \State Assign remaining queries to a query group $\mathcal{G}^q_{N_g}$
\EndIf
\end{algorithmic}
\end{algorithm*}
\begin{table*}[ht!]
\centering
\setlength{\tabcolsep}{5.25pt}
    \begin{tabular}{c|cccccccccc}
    \toprule
        Model & $N_g$ & $k$ & $\lambda_{rel}$ & $\mathcal{R}$ &Optimizer & Training steps & Initial lr & lr decay step & Decayed lr & Batch size \\
    \midrule
        ISG~\cite{khandelwal2022iterative} & 4 & 5 & -0.5 & max & AdamW~\cite{loshchilov2017decoupled} & 150k iters & $10^{-4}$ & 96k'th iter & $10^{-5}$ & 20\\        
        HOTR~\cite{kim2021hotr} & 5 & 4 & -0.5 & max & AdamW & 150k iters & $10^{-4}$ & 96k'th iter &$10^{-5}$  & 20\\        
        GEN~\cite{liao2022gen} & 2 & 5 & -0.5 & max & AdamW & 50 epochs & $10^{-4}$ & 40'th epoch& $10^{-5}$  & 16\\        
    \bottomrule
  \end{tabular}
  \caption{\textbf{Training details and hyperparemeters.}}
  \label{tab:supp_hparams_tab}
\end{table*}

\section{Implementation details.}
For Scene Graph Generation, we implement the proposed SpeaQ on ISG~\cite{khandelwal2022iterative} and HOTR~\cite{kim2021hotr}.
Both architectures adopt a ResNet-101~\cite{he2016deep} backbone and a 6-layer Transformer encoder.
ISG adopts three separate 6-layer decoders for subject, predicate, and object where each decoder takes 300 queries as input, respectively.
HOTR consists of a 6-layer instance decoder and a 12-layer predicate decoder, where each decoder takes 100 and 160 decoder queries as input. 
For Human-Object Interaction Detection, we implement SpeaQ on the smallest model of GEN~\cite{liao2022gen}, $\text{GEN-VLKT}_{s}$.
$\text{GEN-VLKT}_{s}$ adopts ResNet-50 as a backbone and consists of an instance decoder and a predicate decoder, where both consist of 3 decoder layers and take 64 decoder queries as input.
Default hyperparameters for proposed components and for training are presented in Tab.~\ref{tab:supp_hparams_tab}.
Following baselines, a Non-Maximum Suppression (NMS) is applied to remove duplicate detections.
The number of predicate decoder queries $N_q$ is multiplied twice in all three architectures compared to the baselines.
Model weights for the backbone, encoder, and decoder are initialized with the DETR weight pre-trained on the Visual Genome and MS-COCO dataset for Scene Graph Generation models and Human-Object Interaction Detection models, respectively.
All experiments are conducted with 4 NVIDIA RTX 3090 GPUs.

\section{Details about frequency-based predicate groups from previous works.}
In Tab. 8 from the main paper, results under adopting predicate groups split on a frequency-basis from previous works BGNN~\cite{li2021bipartite} and SHA~\cite{dong2022stacked} as $\{\mathcal{G}^p_i\}^{N_g}_{i=1}$ are reported.
In BGNN, predicates are split into three groups ($N_g = 3$) named head, body, and tail by the number of training samples where predicates in each of the three groups have more than 10k samples (\q{head}), between 0.5k and 10k samples (\q{body}), and less than 0.5k samples (\q{tail}), respectively.
In SHA, predicates are split into multiple groups in the way that the number of training samples of the most common predicate in a group does not exceed the pre-defined threshold $\mu$ multiplied by the number of the least common predicate in the same group.
Regarding the reference to \q{SHA} in Tab.~8 of the main paper, we adopt predicate groups constructed under $\mu = 5$.

\section{A running example of quality-aware multi-assignment.}
\begin{figure*}[ht!]
\centering
\includegraphics[width=\textwidth]{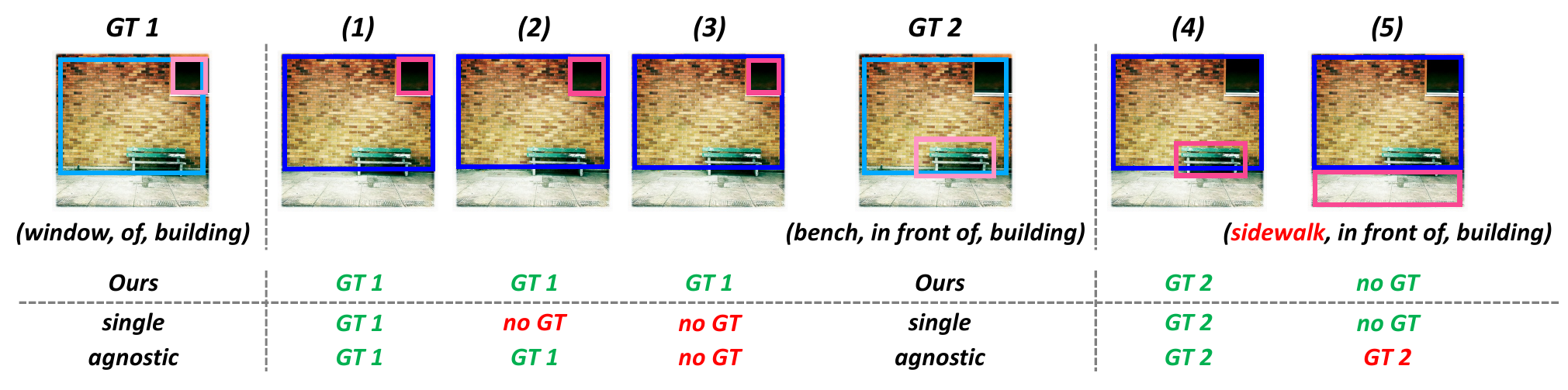}
\caption{
\textbf{Qualitative examples of various assignment strategies.} Bounding boxes and labels of two ground-truths (GT 1, 2) and five prediction results (1-5) are illustrated. Note that a prediction label is only specified in case it differs from the most relevant GT. Ideal assignment results are colored {\color{ForestGreen}green}, while wrong assignment results are colored \textcolor{red}{red}.}
\vspace{-0.25cm}
\label{fig:matching_figure}
\end{figure*}

In this section, we demonstrate the necessity of the proposed `quality-aware' multi-assignment \textbf{(Ours)} over the conventional single assignment \textbf{(single)} and quality-agnostic multi-assignment \textbf{(agnostic)} with the qualitative results in Fig.~\ref{fig:matching_figure}.
To be specific, quality-agnostic multi-assignment denotes that $d_i$, the number of predictions a GT $t_i$ is assigned to, is set equal for every GT, which is two in this example.
With regard to an \q{ideal} assignment, GT 1 should be assigned to predictions 1 to 3, and GT 2 should be assigned to prediction 4 where predicted bounding boxes and classes equal to that of the corresponding GT.
For prediction 5, \q{no relation} should be assigned since both the classification and localization results on the subject largely differ from the GT.
As illustrated in the figure, ours succeeded in finding the ideal assignment.
In contrast, the conventional assignment fails to assign GT 1 to predictions 2 and 3 due to a constraint that a GT can only be assigned to a single GT, and quality-agnostic multi-assignment wrongly assigns GT 2 to prediction 5 due to a constraint that $d_i$ is set equal to every GT.
The qualitative examples demonstrate the importance of adaptively assigning a GT to multiple predictions, since insufficient or wrong training signals may be provided otherwise.

\section{Ratio of \q{no relation} assigned to promising predictions.}
In Tab.~\ref{tab:misassign_rate_tab}, we report the ratio of promising predictions labeled as \q{no relation ($\varnothing$)} to the total number of promising predictions, where a promising prediction is defined as a prediction that is correctly classified and overlaps to the GT in both subject and object with IoU over the certain threshold.
The results show that SpeaQ consistently reduces the ratio, which implies that  abundant positive training signals are provided to promising predictions.

\begin{figure*}[t!]
\centering
\begin{minipage}{0.25\linewidth}
    \includegraphics[width=\linewidth]{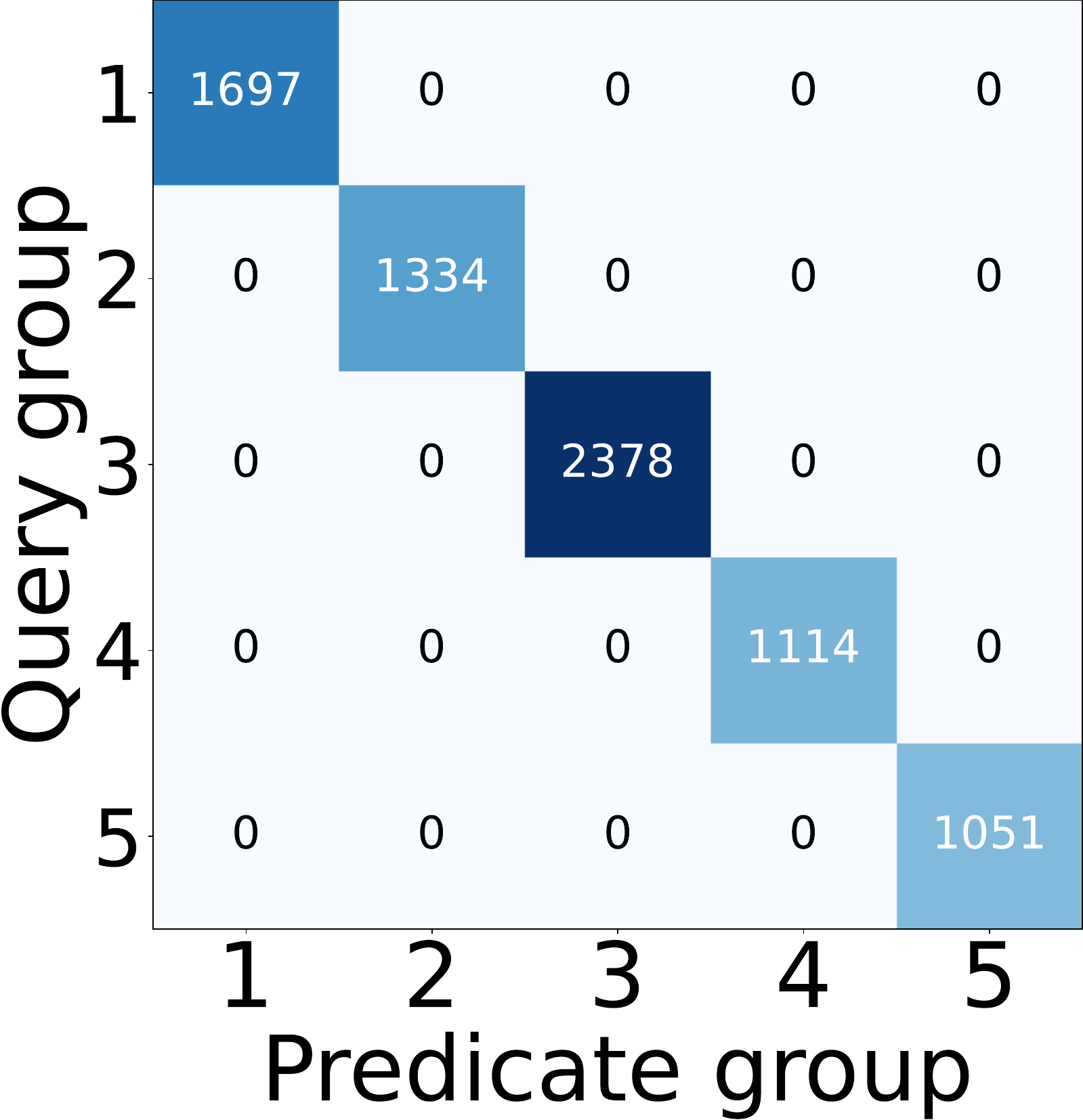}
\vspace{-0.5cm}
\end{minipage}
\caption{ \textbf{Assignment Results between predicate and query groups under SpeaQ.} An element in the $i$-th column and the $j$-th row denotes an average number of predicates in $\mathcal{G}^p_i$ assigned to a query in $\mathcal{G}^q_j$.}

\label{fig:training_signal_per_query}
\end{figure*}

\begin{table}
\centering
\renewcommand{\arraystretch}{0.8}
\setlength{\tabcolsep}{7pt}
    \begin{tabular}{c|ccc}
    \toprule
       Method &  IoU \ensuremath > 0.6 &  IoU \ensuremath > 0.7 & IoU \ensuremath > 0.8 \\
    \midrule
       Baseline &  44.53\% &  42.72\% & 42.61\% \\
       Ours &  \textbf{35.38\%} &  \textbf{33.58\%} & \textbf{33.61\%} \\
    \bottomrule
  \end{tabular}
 \caption{  \textbf{Ratio of promising predictions assigned \q{no relation($\varnothing$)} as a GT.} The lower the percentage, the better.}
\vspace{-0.5cm}
\label{tab:misassign_rate_tab}
\end{table}

\section{Analysis of training signals provided to a query.}
In this section, we validate that a query receives more specialized and abundant training signals under SpeaQ.
First, we provide statistics about the predicate group of a predicate assigned to a query belongs to.
In Fig.~\ref{fig:training_signal_per_query}, a matrix is plotted where an element in the $i$-th column and the $j$-th row denotes an average number of predicates from an $i$'th predicate group assigned to a query in a $j$'th query group.
As shown in the figure, a GT in a specific predicate group is only assigned to a query in the corresponding query group under SpeaQ, given that every entry in the matrix except for diagonal entries is set to zero.
Second, we provide an average number of GTs assigned to a query. 
Under the conventional assignment, 612 samples are assigned to a query on average.
In contrast, 1,540 samples are assigned to a query on average under the SpeaQ, which results in richer training signals being provided to queries.
Considering both statistics provided, we conclude that a query receives more specialized and abundant training signals under SpeaQ.

\begin{figure*}[t!]
\begin{center}
\includegraphics[width=\textwidth]{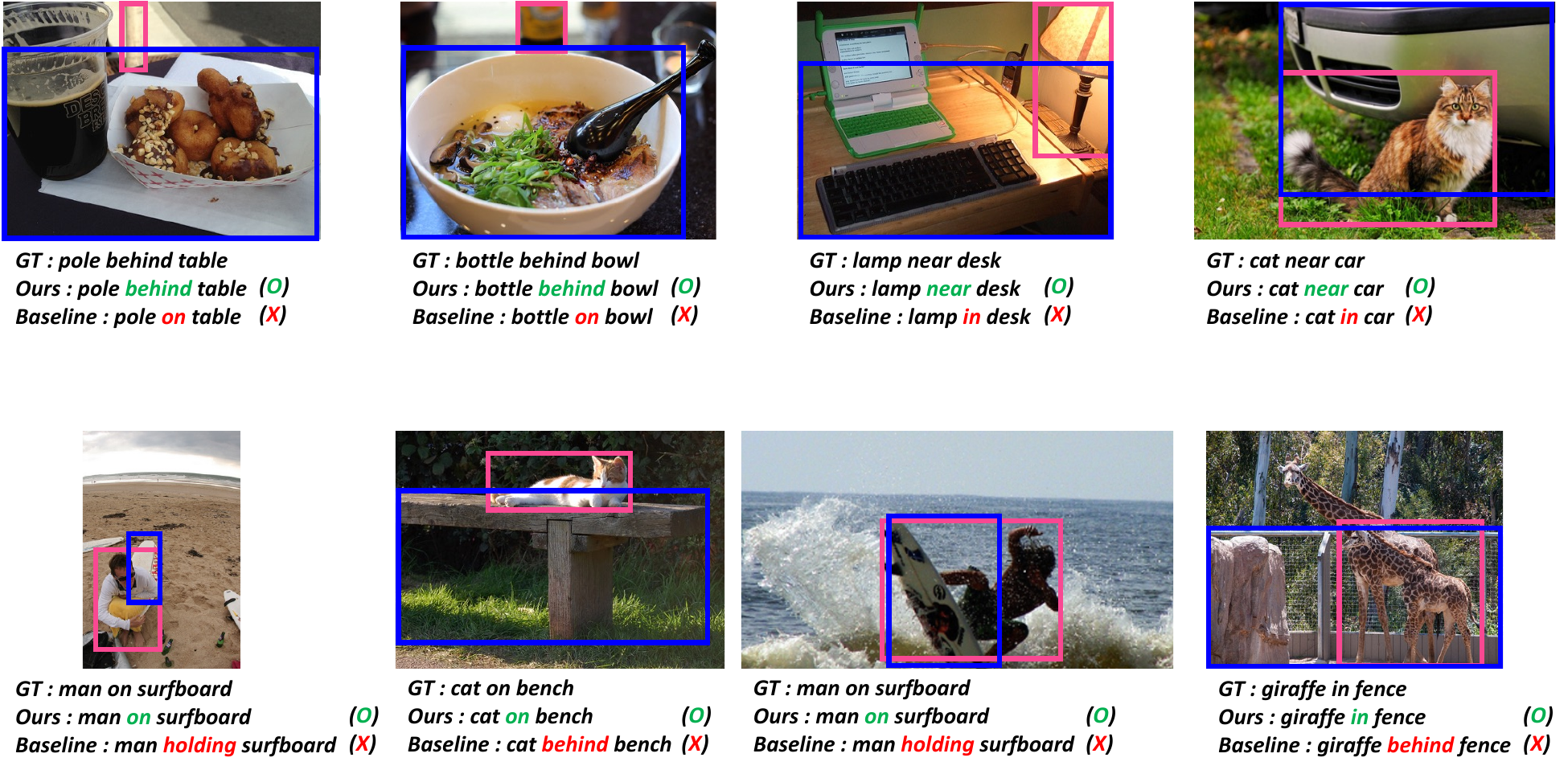}
\end{center}
\vspace{-0.5cm}
\caption{
\textbf{Further qualitative results on Visual Genome dataset.} Prediction results of the baseline and the model trained with SpeaQ are visualized. Predicates classified correctly are marked {\color{ForestGreen}{green}}, while predicates that are misclassified are marked \textcolor{red}{red}.}
\label{fig:supp_qual_figure}
\end{figure*}

\begin{figure*}[t!]
\begin{center}
\includegraphics[width=\textwidth]{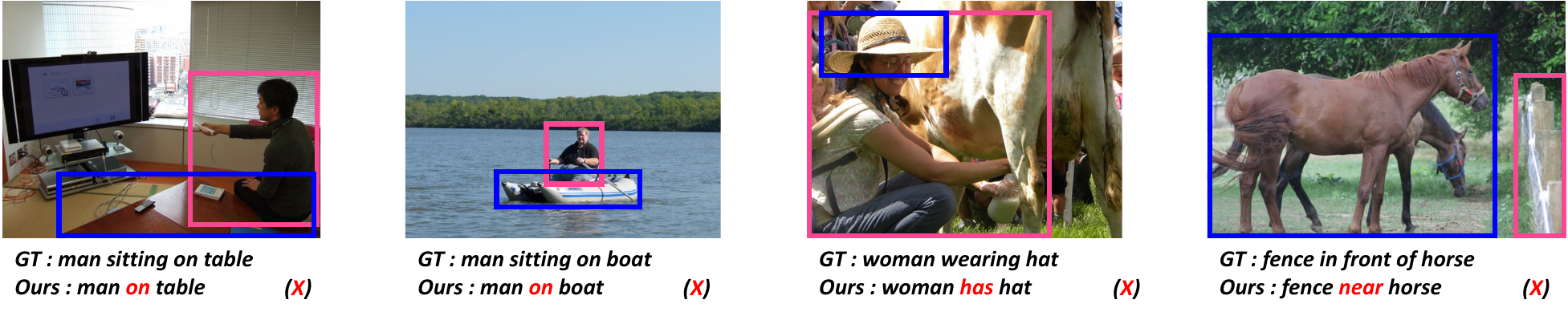}
\vspace{-0.5cm}
\end{center}
\caption{
\textbf{Failure cases on Visual Genome dataset.} Prediction results of the model trained with SpeaQ are visualized.}
\label{fig:supp_qual_failure}
\vspace{-0.5cm}
\end{figure*}

\section{Further qualitative examples.}
In Fig.~\ref{fig:supp_qual_figure}, further qualitative examples are provided.
Examples demonstrate that the model trained with SpeaQ succeeds in correctly detecting challenging samples that require both semantic and visual understanding compared to the baseline.

\section{Failure cases.}
In Fig.~\ref{fig:supp_qual_failure}, we provide qualitative results where the model trained with SpeaQ fails.
As shown in the figure, although the predicted label is considered incorrect based on the GT annotations, some predictions are not completely wrong due to the ambiguity of the GT or language.
Therefore, we suggest developing a more accurate evaluation metric or annotations addressing the ambiguity of GTs may be an interesting direction for future research.
\newpage

\end{document}